\pdfoutput=1

\documentclass[11pt]{article}

\usepackage[final]{acl}

\usepackage{times}
\usepackage{latexsym}

\usepackage[T1]{fontenc}

\usepackage[utf8]{inputenc}

\usepackage{microtype}

\usepackage{inconsolata}

\usepackage{graphicx}
\usepackage{float}
\usepackage{amsfonts}
\usepackage{amsmath}
\usepackage{hyperref}
\usepackage{url}
\usepackage{graphicx}
\usepackage{multirow}
\usepackage{caption}
\usepackage{subcaption}
\usepackage{booktabs}
\usepackage{nicefrac}
\usepackage{microtype}
\usepackage{xcolor}
\usepackage{multirow}
\usepackage{epsfig}
\usepackage{amsmath}
\usepackage{diagbox}
\usepackage{float}
\usepackage{pifont}
\usepackage{color}
\usepackage{xspace}
\usepackage{caption}
\usepackage{wrapfig}
\usepackage{varwidth}

\newcolumntype{M}{>{\begin{varwidth}{4cm}}l<{\end{varwidth}}}

\newcommand*{\rowstyle}[1]{
  \gdef\@rowstyle{#1}%
  \@rowstyle\ignorespaces%
}
\newcolumntype{=}{
  >{\gdef\@rowstyle{}}%
}
\newcolumntype{+}{
  >{\@rowstyle}%
}
\definecolor{revcolor}{rgb}{0,0,0}

\title{HyperBERT: Mixing Hypergraph-Aware Layers with Language Models for Node Classification on Text-Attributed Hypergraphs}

\author{Adrián Bazaga, Pietro Liò, Gos Micklem \\
        University of Cambridge, Cambridge, United Kingdom \\
        \{$ar989,pl219,gm263$\}@cam.ac.uk}

\begin{document}
\maketitle
\begin{abstract}
Hypergraphs are characterized by complex topological structure, representing higher-order interactions among multiple entities through hyperedges. Lately, hypergraph-based deep learning methods to learn informative data representations for the problem of node classification on text-attributed hypergraphs have garnered increasing research attention. However, existing methods struggle to simultaneously capture the full extent of hypergraph structural information and the rich linguistic attributes inherent in the nodes attributes, which largely hampers their effectiveness and generalizability. To overcome these challenges, we explore ways to further augment a pretrained BERT model with specialized hypergraph-aware layers for the task of node classification. Such layers introduce higher-order structural inductive bias into the language model, thus improving the model's capacity to harness both higher-order context information from the hypergraph structure and semantic information present in text. In this paper, we propose a new architecture, HyperBERT, a mixed text-hypergraph model which simultaneously models hypergraph relational structure while maintaining the high-quality text encoding capabilities of a pre-trained BERT. Notably, HyperBERT presents results that achieve a new state-of-the-art on five challenging text-attributed hypergraph node classification benchmarks.
\end{abstract}

\section{Introduction}

In recent years, the study of hypergraphs has gained significant attention in the field of network analysis \cite{BAI2021107637, 9264674, HOU2022108526}. Unlike traditional graphs, where edges represent pairwise relationships, hypergraphs are a generalization of graphs where an edge can join any number of nodes, allowing for more complex interactions by connecting multiple nodes through hyperedges. This characteristic makes hypergraphs an ideal framework for representing complex systems such as social networks, biological networks, and collaboration networks, where relationships often involve sets of entities.

Machine learning on hypergraphs \cite{9782536, BAI2021107637}, has been shown to be an effective tool for studying complex graph-based data structures. In particular, when compared with a standard graph, a hypergraph is more capable and flexible in modeling high-order relations since one hyperedge in a hypergraph can connect more than two nodes. As a result, hypergraph learning has recently gained increasing attention and has been applied to many research fields, such as computer vision \cite{Feng_You_Zhang_Ji_Gao_2019,HOU2023109035} and recommendation systems \cite{10.1145/3544105,10018469}.

In many real-world complex systems, hypergraph data is accompanied by node attribute information \cite{nakajima2024inferring,Failla2023,SMANIOTTO2021143,NEURIPS2021_f18a6d1c}. Some examples of node
attributes include the age, ethnicity, and gender of individuals in social networks \cite{Contisciani2020}, affiliations of authors in collaboration networks \cite{Pan2012}, and article abstract texts in co-citation graphs \cite{hu2021open}. Previous studies have demonstrated that node attribute data potentially enhances the learning of community structure in networks \cite{CHUNAEV2020100286, doi:10.1126/sciadv.abn7558}.

A particular instance of attributed hypergraphs are Text-Attributed Hypergraphs (TAHGs) \cite{nakajima2024inferring,Failla2023,SMANIOTTO2021143,NEURIPS2021_f18a6d1c}, which have been widely used for modeling a variety of real-world applications, such as co-authoring in collaboration networks \cite{doi:10.1073/pnas.98.2.404,Patania2017}, information retrieval \cite{cohan-etal-2020-specter, NEURIPS2021_f18a6d1c}, product recommendation \cite{10.1145/3442381.3449842}, and many others.

Given the prevalence of TAHGs \cite{nakajima2024inferring,Failla2023,SMANIOTTO2021143,NEURIPS2021_f18a6d1c}, we have explored how to effectively handle these graphs, with a focus on node classification tasks. Such tasks aims at predicting the category or label of a node based on its textual attributes and the structure of the hypergraph. This problem is pivotal in various applications such as predicting protein functions in biological networks \cite{10.1093/bioinformatics/btaa768}, identifying roles in social networks \cite{doi:10.1073/pnas.98.2.404,Patania2017}, and classifying documents in citation networks \cite{hu2021open}. Inherently, TAHGs provide both node attribute and graph structural information. Thus, it is critical to effectively capture both while modeling their interrelated correlation for effective representation learning.

Node classification in text-attributed hypergraphs presents unique challenges that are not encountered in traditional graph-based approaches \cite{10.1145/3589261,WU2021107185,Failla2023}. In this context, there are two key hurdles. First, since hyperedges can connect any set of nodes, capturing such complex interaction patterns becomes increasingly difficult using conventional graph-based methods. Second, effectively exploiting the textual information to learn informative node representations is complex. Recent advances in the field of NLP have led to the emergence of language models (LMs) with strong capabilities for semantic understanding of text \cite{Devlin2019BERTPO,raffel_exploring_2020, NEURIPS2020_1457c0d6, radford2019language, radford2018improving}. However, the integration of hypergraph structural information directly into language models is still an open and under-researched topic.

To address the above limitations, we propose a novel architecture, HyperBERT, which mixes hypergraph-aware layers into language models to simultaneously exploit hypergraph topology and textual representations for the node classification task on text-attributed hypergraphs. In particular, HyperBERT extends the original BERT architecture with hypergraph-specific inductive biases. To do so, we design specially-crafted hypergraph-aware layers into the BERT encoder, effectively mixing hypergraph structure with text semantic embeddings for computing node representations. To accomplish the text-hypergraph alignment in the feature space, we propose a novel self-supervised loss for pretraining HyperBERT and effectively aligning semantic and hypergraph feature spaces. HyperBERT achieves state-of-the-art performance across five widely used hypergraph benchmarks. Our code is available at \href{https://github.com/AdrianBZG/HyperBERT}{https://github.com/AdrianBZG/HyperBERT}.

\section{Related Work}

Our work is related to two lines of research: text-attributed graph learning-based methods and language model-based approaches on text-attributed graphs \cite{zhao2023learning,NEURIPS2021_f18a6d1c,kuang-etal-2023-unleashing}. In this section, we review briefly the relevant literature for such categories.

Given the recent success of graph representation learning, numerous research works have been proposed for various tasks including node classification \cite{Bhagat2011NodeCI} or link prediction \cite{10.5555/3327345.3327423}. Graph Neural Networks (GNNs) are recognized as the de facto standard technique for modeling graph data. These methods (e.g. GCN; \citet{kipf2017semisupervised}, GAT; \citet{veličković2018graph}, GraphSAGE; \citet{hamilton2017inductive} or GIN; \citet{xu2018how}) learn effective mechanisms such that information between the nodes is aggregated for expressive graph representations.  In the case of TAGs, a cascade architecture \cite{hamilton2017inductive} is typically adopted. In such setting node features are encoded independently using text modeling techniques (e.g. Bag-of-Words \cite{Zhang2010}, skip-gram \cite{NIPS2013_9aa42b31}, or more recently language models such as BERT \cite{Devlin2019BERTPO}), and subsequently aggregated via the message passing mechanism to produce the final node representations. However, standard GNN-based approaches are only applicable to general graphs with binary relations. 

To alleviate this issue recent research has focused on developing effective methodologies for attributed hypergraph learning in order to capture n-ary or high-order interactions between nodes \cite{10.1145/3494567,10.1145/3308558.3313635,10.1145/3544105,wang2020next}. For instance, \citet{zhang2018} proposes an inductive multi-hypergraph learning algorithm, which learns an optimal hypergraph embedding with good performance on the task of multi-view 3D object classification. More recently, several hypergraph-based methods have been proposed for the task of node classification \cite{10.5555/3454287.3454422, BAI2021107637}. In particular, \cite{Feng_You_Zhang_Ji_Gao_2019} propose the Hypergraph Neural Network (HGNN) architecture, which encodes high-order node correlations using a hypergraph structure and generalizes the convolution operations to the hypergraph learning process based on hypergraph Laplacian and truncated Chebyshev polynomials. \citet{BAI2021107637} propose to supplement the family of graph neural networks with two end-to-end trainable operations, i.e. hypergraph convolution and hypergraph attention (HCHA). \citet{10.5555/3454287.3454422} propose HyperGCN, model to enhance the hypergraph Laplacian with additional weighted pair-wise edges. \citet{kim2024hypeboy} propose HypeBoy, a hypergraph self-supervised learning method for node classification. However, HypeBoy primarily focuses on hypergraphs without specifically addressing the complexities introduced by nodes with textual attributes.

Given the success of LMs in solving semantic representation learning tasks, recent works \cite{zhang2021greaselm} have proposed to utilise LMs for TAGs, as they can transform the nodes’ textual attributes into semantically rich node embeddings. However, node embeddings provided by LMs often only contain information from the textual attributes, overlooking the topological structure present in the TAGs. To overcome this limitation, multiple works integrate LM-based embeddings with graph learning methods \cite{jin2023edgeformers, jin2023heterformer}, with the goal of generating node embeddings that are tailored for specific TAG tasks.

For instance, \citet{NEURIPS2021_f18a6d1c} propose Graphormer, which fuses LM-based text encoding and graph aggregation into an iterative workflow. GIANT \cite{Chien2021NodeFE} introduces a novel neighborhood prediction task to fine-tune a XR-Transformer \cite{zhang2021fast}, and feeds node embeddings as obtained by the LM into GNNs for downstream tasks. TAPE \cite{he2023harnessing} presents a method that uses LMs to generate prediction text and explanation text, which serve as augmented text data to subsequently feed into GNNs. LM-GNN \cite{ioannidis2022efficient} introduces graph-aware pre-training to adapt the LM on the given graph before fine-tuning the entire model, demonstrating significant performance results. SimTeG \cite{duan2023simteg} finds that first training the LMs on the downstream task and then freezing the LMs and training the GNNs can result in improved performance. K-BERT \cite{Liu_Zhou_Zhao_Wang_Ju_Deng_Wang_2020} leverages a knowledge layer to inject relevant triplets from a knowledge graph into the input sentence and transform it into a knowledge-rich sentence tree. KG-BERT \cite{yao2019kgbert} treats triples in knowledge graphs as textual sequences. They propose to model entity and relation descriptions of a triple as input, and compute the scoring function of the triple with the KG-BERT language model. Despite these advances, previous works do not take into account the problem where both text attributes and higher-order interactions exist between the entities, as is the case with text-attributed hypergraph node classification.

\section{Preliminaries}

In this paper, we focus on learning representations for nodes in Text-Attributed Hypergraphs (TAHG), where we take node classification as the downstream task. Therefore, before describing the details of our proposed method, we start with presenting a few basic concepts, including the formal definition of TAHGs and how language models can be used for node classification in TAHGs. Then, we introduce the formulation of the problem our method is tackling. 

\begin{figure}[h]
\centering
\includegraphics[width=1.0\columnwidth]{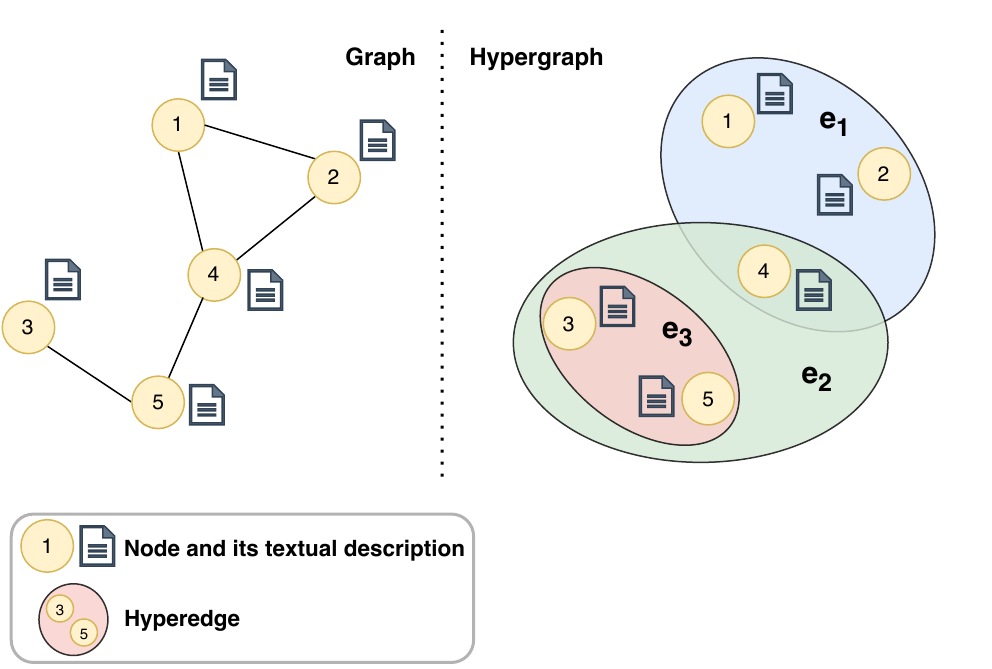}
\caption{An illustration of a standard text-attributed graph (left) where nodes are connected between each other using binary relations, and a text-attributed hypergraph (right) where nodes are related with high-order connections (hyperedges). In both cases, each node is attributed with a textual description, such as paper abstract in the case for co-citation hypergraphs.}
\label{fig:figure1}
\end{figure}

\subsection{Text-Attributed Hypergraphs (TAHG)}

Formally, a hypergraph $\mathcal{G}$ = ($\mathcal{V}$, $\mathcal{E}$, $\mathcal{S}^{V}$) is defined by a node set $\mathcal{V}$ and a hyperedge set $\mathcal{E}$. Each hyperedge $e_{j}$ $\in$ $\mathcal{E}$ is a non-empty set of nodes. Each node $v_{i}$ $\in$ $\mathcal{V}$ is associated with a sequential text feature, $s_{i}$ $\in$ $\mathcal{S}^{V}$, and the corresponding label $y_{i}$. Each label has a real label text $c$ from the set of all label texts $C$ (e.g. 'Artificial Intelligence' or 'Supervised Learning' for the case of paper categories in a co-citation hypergraph). For the sake of clarity, Figure \ref{fig:figure1} illustrates the difference between standard text-attributed graphs and text-attributed hypergraphs.

\subsection{Pretrained Language Models (PLM) for Node Classification}

In the context of TAGs, LMs can be employed to encode the text attributes associated with each node and learn a representation that captures the semantic meaning of the text. Let $s_{i}$ $\in$ $\mathcal{S}^{V}$ denote the text attributes of node $i$, and LM be a pre-trained network, such as BERT \cite{Devlin2019BERTPO}. Then, the text attributes of node $i$ can be encoded by applying BERT to $s_{i}$. The output of the LM is denoted as $\mathbf{z_{i}}$ $\in$ $\mathbb{R}^{d}$, where $d$ is the dimensionality of the output feature vector. Then, to perform node classification, the output is typically fed into a fully connected layer with a softmax function for class prediction. The goal is to learn a function that maps the encoded text attributes to the corresponding node labels.

\subsection{Problem Formulation}

Given a text-attributed hypergraph $\mathcal{G}$ = ($\mathcal{V}$, $\mathcal{E}$, $\mathcal{S}^{V}$), with each node associated with a textual description $\mathcal{S}$ = \{$s_1, ..., s_{|\mathcal{S}|}$\}, where $|\mathcal{S}|$ is the number of tokens in the textual description $\mathcal{S}$, and a partial set of node labels $C$ $\subset$ $\mathcal{V}$, the goal is to predict the labels of the remaining nodes $P$ = $\mathcal{V}$ $\setminus$ $C$.

\section{Methodology}

\begin{figure*}[th!]
\centering
\includegraphics[width=2.0\columnwidth]{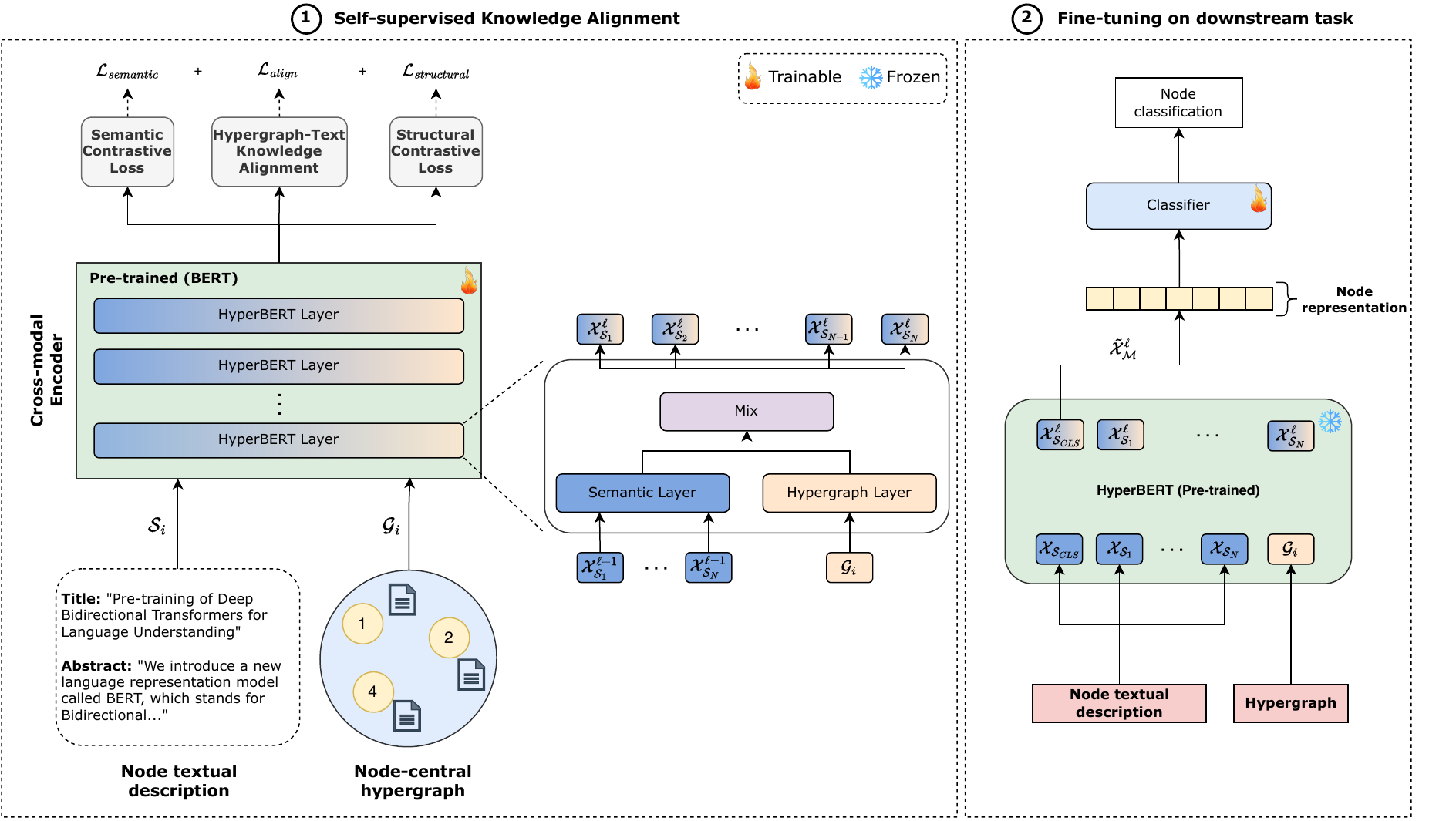}
\caption{High-level overview of our proposed HyperBERT model. The model mixes hypergraph-aware layers into BERT to simultaneously exploit hypergraph topology and text semantics for node classification tasks on TAHGs. To accomplish the text-hypergraph alignment in the feature space, HyperBERT employs a novel self-supervised loss for pretraining that effectively aligns semantic and hypergraph feature spaces. After pretraining the model, it can be fine-tuned on a variety of downstream hypergraph node-level tasks such as node classification.}
\label{fig:hyperbert1}
\end{figure*}

In this section we present our proposed model, HyperBERT, as illustrated in Figure \ref{fig:hyperbert1}. First, we begin with the details of the HyperBERT layer design. Then, we detail our proposed pre-training methodology, followed by details of the different components of our knowledge alignment loss function. Finally, we describe how fine-tuning is performed for a downstream node classification task.

\subsection{HyperBERT Layer}

The HyperBERT layer explicitly mixes semantic information obtained from each BERT transformer block with structural information obtained with a Hypergraph Neural Network (HGNN) block.

\paragraph{Semantic Representation.} The semantic representations of node textual attributes are encoded by a BERT \cite{Devlin2019BERTPO} encoder. Specifically, BERT consists of alternating layers of multi-head self-attention (MHA) and Fully-connected Forward Network (FFN) blocks, as in the original Transformer \cite{vaswani_attention_2017} architecture. Before every block, Layer Normalisation (LN) is applied, and after every block, a residual connection is added. More formally, in the $\ell^{th}$ encoder layer, the hidden states are represented as $\mathcal{X}^{\ell}_\mathcal{S} = \{x^{\ell}_{1}, \ldots, x^{\ell}_{N}\}$, where \textit{N} is the maximum length of the inputs. First, a MHA block maps $\mathcal{X}$ into a query matrix $\mathbf{Q}$ $\in$ $\mathbb{R}^{n\times d_{k}}$, key matrix $\mathbf{K}$ $\in$ $\mathbb{R}^{n\times d_{k}}$ and value matrix $\mathbf{V}$ $\in$ $\mathbb{R}^{n\times d_{v}}$, where \textit{m} is the number of query vectors, and \textit{n} the number of key/value vectors. Then, an attention vector is calculated as follows

\begin{equation}
\begin{aligned}
\mathbf{Attn(Q, K, V)} {=} & \mathbf{softmax(A)} \mathbf{V},\\
\mathbf{A} {=} & \frac{\mathbf{Q} \mathbf{K^T}}{\mathbf{\sqrt{d_{k}}}}
\end{aligned}
\label{eq:attn_formula}
\end{equation}

In practice, the MHA block calculates the self-attention over \textit{h} heads, where each head \textit{i} is independently parametrized by $\mathbf{W^{Q}_{i}}$ $\in$ $\mathbb{R}^{d_{m}\times d_{k}}$, $\mathbf{W^{K}_{i}}$ $\in$ $\mathbb{R}^{d_{m}\times d_{k}}$ and $\mathbf{W^{V}_{i}}$ $\in$ $\mathbb{R}^{d_{m}\times d_{v}}$, mapping the input embeddings $\mathcal{X}$ into queries and key-value pairs. Then, the attention for each head is calculated and concatenated, as follows

\begin{equation}
\begin{aligned}
\mathbf{Head_{i}} {=} & \mathbf{Attn(Q W^{Q}_{i}, K W^{K}_{i}, V W^{V}_{i})}\\
\mathbf{MHA(\mathcal{X}^{\ell}_\mathcal{S})} {=} & \mathbf{Concat(Head_{1}, \ldots, \textnormal{Head}_{h}) W^{U}}\\
\mathbf{\bar{\mathcal{X}}^{\ell}_\mathcal{S}} {=} & \mathbf{MHA(\mathcal{X}^{\ell}_\mathcal{S})}
\end{aligned}
\end{equation}

where $\mathbf{W^{U}}$ $\mathbf{\in}$ $\mathbb{R}^{d^{h}_{m}\times d_{m}}$ is a trainable parameter matrix. Next, to acquire the semantic hidden states of the input, a FFN block is applied, as follows

\begin{equation}
\mathbf{FFN(\bar{\mathcal{X}}^{\ell}_\mathcal{S})} = \mathbf{max(0, \bar{\mathcal{X}}^{\ell}_\mathcal{S} W_{1} + b_{1}) W_{2} + b_{2}}
\end{equation}

where $\mathbf{W_{1}}$ $\in$ $\mathbb{R}^{d_{m}\times d_{ff}}$ and $\mathbf{W_{2}}$ $\in$ $\mathbb{R}^{d_{ff}\times d_{m}}$ are linear weight matrices. Finally, layer normalisation and residual connection are applied as follows

\begin{equation}
\mathbf{\tilde{\mathcal{X}}^{\ell}_\mathcal{S} = \mathbf{LayerNorm}(\bar{\mathbf{\mathcal{X}}}^{\ell}_\mathcal{S} + \mathbf{FFN}(\bar{\mathbf{\mathcal{X}}}^{\ell}_\mathcal{S}))}
\end{equation}

Therefore, after the \textit{L} encoder layers, we obtain the nodes' textual description semantic representations as the pooled feature tensor of the [CLS] token, denoted as $\tilde{\mathcal{X}}^{\ell}_\mathcal{S}$.

\paragraph{Hypergraph Structural Representation.} In each HyperBERT layer, structural representations are produced through a Hypergraph Neural Network (HGNN) \cite{Feng_You_Zhang_Ji_Gao_2019} over the node-centered context hypergraph $\mathcal{G}_{i}$, defined as the subhypergraph containing only the hyperedges which node $i$ belongs to, and its incidence matrix $H_{i}$. Formally, we formulate the HGNN block to obtain hypergraph node embeddings as follows

\begin{equation}
\tilde{\mathcal{X}}^{\ell}_\mathcal{G} = \sigma(\mathbf{D}^{-1}\mathbf{HW}\mathbf{B}^{-1}\mathbf{H}^\mathrm{T}\mathbf{X}^{(\ell-1)}\mathbf{\Psi}),
\end{equation}

where $\mathbf{D}$ and $\mathbf{B}$ are the degree matrices of the vertex and hyperedge in the hypergraph, respectively. $\sigma(\cdot)$ is a non-linear activation function such as ReLU \cite{Maas2013ReLU}. $\mathbf{\Psi} \in \mathbb{R}^{d_{\ell}\times d_{(\ell-1)}}$ is a learnable weight matrix between the $\ell^{th}$ and $(\ell-1)^{th}$ layer. Consequently, after \textit{L} hypergraph layers, we obtain the nodes' structural representations as $\tilde{\mathcal{X}}^{\ell}_\mathcal{G}$.

\paragraph{Text-Hypergraph Joint Representation.} After computing representations from both semantic and hypergraph structural spaces, the $l^{th}$ HyperBERT layer computes a mixed text-hypergraph representation as follows

\begin{equation}
\tilde{\mathcal{X}}^{\ell}_\mathcal{M} = \tilde{\mathcal{X}}^{\ell}_\mathcal{S} + \tilde{\mathcal{X}}^{\ell}_\mathcal{G},
\end{equation}

where $\tilde{\mathcal{X}}^{\ell}_\mathcal{M}$ contains a mixture of semantic and hypergraph structural information to enable information flow between both modalities. Specifically, to combine structural and semantic embeddings, we add the hypergraph embedding to the embedding of each of the tokens in $\tilde{\mathcal{X}}^{\ell}_\mathcal{S}$. Therefore, $\tilde{\mathcal{X}}^{\ell}_\mathcal{M}$ effectively incorporates both the semantic information from text attributes and the structural context from the hypergraph. This is crucial for capturing the complex interplay between textual content and graph topology.

\subsection{Hypergraph-Aware Pretraining Task}

In order to improve the learning capability of HyperBERT without using any human-annotated labels, we propose a novel hypergraph-aware knowledge alignment pretraining algorithm based on a contrastive learning loss. The proposed hypergraph-aware pretraining task exploits inherent hypergraph knowledge from the TAHG and is applied to both the semantic and structural representations.

\paragraph{Semantic Contrastive Loss.} Based on the semantic representation of node $i$, $\tilde{\mathcal{X}}_{\mathcal{S}_{i}}$, the other nodes belonging to its same set of hyperedges, $\mathcal{N}(i)$, and the node ${i}$ excluded mini-batch instances, $\mathcal{B}(i)$, the semantic contrastive loss objective can be defined as follows: 

\begin{equation}
 \mathcal{L}_{\texttt{semantic}}{=}\frac{-1}{|\mathcal{N}(i)|} \sum_{p\in \mathcal{N}(i)}{\log}  \frac{ 
 e^{(\tilde{\mathcal{X}}_{\mathcal{S}_{i}} \cdot \tilde{\mathcal{X}}_{\mathcal{S}_{p}}/\tau)}}{\sum\limits_{j\in \mathcal{B}(i)} e^{(\tilde{\mathcal{X}}_{\mathcal{S}_{i}} \cdot  \tilde{\mathcal{X}}_{\mathcal{S}_{j}}/\tau)}},
\end{equation}

where $\tau$ denotes the temperature and $\text{e}(\cdot)$ represents the exponential function. Here for node $i$, we consider its semantic representation $\tilde{\mathcal{X}}_{\mathcal{S}_{i}}$ as the query instance. The positive instances are the representations of node $i$'s hyperedge members $\{\tilde{\mathcal{X}}_{\mathcal{S}_{p}} \mid p\in \mathcal{N}(i)\}$. Meanwhile, the negative instances are the representations of other text nodes excluding $i$ within the same mini-batch $\{\tilde{\mathcal{X}}_{\mathcal{S}_{j}} \mid j \in \mathcal{B}(i)\}$.

\paragraph{Structural Contrastive Loss.} Similar to the semantic representation, we also apply our contrastive learning algorithm to the hypergraph structural feature space representation of node $i$, denoted as $\tilde{\mathcal{X}}_{\mathcal{G}_{i}}$. Formally, the structural contrastive loss is formulated as follows:

\begin{equation}
    \mathcal{L}_{structural}{=}\frac{-1}{|\mathcal{N}(i)|}
 \sum_{p\in \mathcal{N}(i)}{\log} \frac{ e^{(\tilde{\mathcal{X}}_{\mathcal{G}_{i}} \cdot  \tilde{\mathcal{X}}_{\mathcal{G}_{p}}/\tau)}}{\sum\limits_{j\in \mathcal{B}(i)}e^{(\tilde{\mathcal{X}}_{\mathcal{G}_{i}} \cdot \tilde{\mathcal{X}}_{\mathcal{G}_{j}}/\tau)}},
\end{equation}

where $\tilde{\mathcal{X}}_{\mathcal{G}_{i}}$ is the query instance. The positive instances are $\{\tilde{\mathcal{X}}_{\mathcal{G}_{p}} \mid p \in \mathcal{N}(i)\}$ and the negative instances are $\{\tilde{\mathcal{X}}_{\mathcal{G}_{j}} \mid j \in \mathcal{B}(i)\}$.

Therefore, our semantic and structural contrastive learning losses ensure nodes belonging to the same hyperedges to share similar representations, which inherently elicits informative hypergraph knowledge during the learning process.

\paragraph{Hypergraph-Text Knowledge Alignment.} In this work, our goal is to learn expressive representations that simultaneously encode informative textual semantics within each text node, as well as structural information among hyperedges. However, individually performing contrastive learning on either the semantic or structural spaces is not enough due to the lack of information exchange between both modalities. To better align the knowledge captured by the semantic and hypergraph structural layers, we propose a hypergraph-text knowledge alignment algorithm for TAHGs. 

In particular, for each node $i$, based on its semantic and structural representations, $\tilde{\mathcal{X}}_{\mathcal{S}_{i}}$ and $\tilde{\mathcal{X}}_{\mathcal{G}_{i}}$, respectively, we formulate the objective of hypergraph-text knowledge alignment loss, $\mathcal{L}_{align}$, as follows:

\begin{equation}
    \begin{aligned}
        \mathcal{L}_{align}{=} & \frac{-1}{|\mathcal{N}(i)|} \sum_{p\in \mathcal{N}(i)}  \biggr(\mathcal{L}_{align_{1}}  + \mathcal{L}_{align_{2}} \biggr) / 2,\\
\mathcal{L}_{align_{1}}{=} & \log \frac{e^{(\tilde{\mathcal{X}}_{\mathcal{G}_{i}} \cdot \tilde{\mathcal{X}}_{\mathcal{S}_{p}}/\tau)}}{\sum\limits_{j\in \mathcal{B} (i)}e^{(\tilde{\mathcal{X}}_{\mathcal{G}_{i}} \cdot \tilde{\mathcal{X}}_{\mathcal{S}_{j}}/\tau)}},\\
\mathcal{L}_{align_{2}}{=} & \log \frac{e^{(\tilde{\mathcal{X}}_{\mathcal{S}_{i}} \cdot  \tilde{\mathcal{X}}_{\mathcal{G}_{p}}/\tau)}}{\sum\limits_{j\in \mathcal{B}(i)} e^{(\tilde{\mathcal{X}}_{\mathcal{S}_{i}} \cdot  \tilde{\mathcal{X}}_{\mathcal{G}_{j}}/\tau)}}
\end{aligned}
\end{equation}

Given the above formulation, in the first component of the knowledge alignment loss function, $\mathcal{L}_{align_{1}}$, we treat the structural representation of node $i$, $\tilde{\mathcal{X}}_{\mathcal{G}_{i}}$, as the query, then construct the positive and negative instances based on the semantic representations. Specifically, the positive instances include both the representation of node $i$ as well as the representations of $i$'s same hyperedge nodes (i.e., $\{\tilde{\mathcal{X}}_{\mathcal{S}_{p}} \mid p \in \widetilde{\mathcal{N}}(i)\}$), and the negative instances are the representations of other instances within the same mini-batch $\{\tilde{\mathcal{X}}_{\mathcal{S}_{j}} \mid j \in \mathcal{B}(i)\}$. In the second component of the knowledge alignment loss, $\mathcal{L}_{align_{2}}$, we consider the semantic representation $\tilde{\mathcal{X}}_{\mathcal{S}_{i}}$ as the query and construct its corresponding positive and negative instances in the parallel way as for the first loss component. Consequently, the proposed $\mathcal{L}_{align}$ alignment loss encourages the representations of the same node learned across the two separate feature spaces, semantic and structural, to be pulled together in the embedding space.

\paragraph{Learning Objective.} In order to learn our hypergraph-aware language model, we jointly optimize the proposed semantic, structural and hypergraph-text knowledge alignment losses. Specifically, the overall training loss objective is formulated as follows:

\begin{equation}
    \mathcal{L}{=} \lambda_{1} \mathcal{L}_{semantic} + \lambda_{2} \mathcal{L}_{structural} + \lambda_{3} \mathcal{L}_{align},
\end{equation}

where $\lambda_{1}$, $\lambda_{2}$, $\lambda_{3}$ $\in$ $\mathbb{R}$. In practice, we found that the values of $\lambda_{1}$ = $\lambda_{2}$ = $\lambda_{3}$ = 1 led to good experimental results.

\subsection{Fine-Tuning in Downstream Tasks}

Once the pretraining is finished, we freeze the parameters of HyperBERT and use it to compute representations of each text node. In particular, during fine-tuning HyperBERT computes the text-hypergraph joint representation, $\tilde{\mathcal{X}}^{\ell}_\mathcal{M}$, capturing an aligned representation of the node's textual attributes and its topological structure in the hypergraph. Furthermore, this representation is used to fine-tune separate models on different downstream tasks. Specifically, we train a linear classifier from the node features to class labels for node classification using the cross entropy loss.

\begin{table*}[th!]
    \renewcommand\arraystretch{1.}
    \centering
    \small
    \caption{Performance of HyperBERT on a variety of hypergraph node classification benchmarks. Cell values indicate mean and standard deviation accuracy calculated over 10 runs. Underscoring depicts the best performance for each dataset across previous methods. Bold indicates the best overall method for each dataset.}
    \setlength{\tabcolsep}{3mm}{
    \small
    \resizebox{1.\linewidth}{!}{
    \begin{tabular}{=c|+c+c|+c+c|+c}
        \hline
        \multirow{2}{*}{\textbf{Method}} & \multicolumn{2}{c|}{\textbf{Co-citation}} & \multicolumn{2}{c|}{\textbf{Co-authorship}} & \multicolumn{1}{c}{\textbf{Movies}} \\
        \cline{2-6}
        & \textbf{Cora} & \textbf{Pubmed} & \textbf{DBLP-A} & \textbf{Cora-CA} & \textbf{IMDB}  \\
        \hline
         HGNN \citep{Feng_You_Zhang_Ji_Gao_2019} & 50.0 \scriptsize{(7.2)} & 72.9 \scriptsize{(5.0)} & 67.1 \scriptsize{(6.0)} & 50.2 \scriptsize{(5.7)} & 42.2 \scriptsize{(2.9)} \\
         HyperGCN \citep{10.5555/3454287.3454422} & 33.1 \scriptsize{(10.2)} & 63.5 \scriptsize{(14.4)} & 68.2 \scriptsize{(14.4)} & 50.2 \scriptsize{(5.7)} & 37.9 \scriptsize{(4.5)} \\
         HNHN \citep{HNHN2020} & 50.0 \scriptsize{(7.9)} & 72.1 \scriptsize{(5.4)} & 62.6 \scriptsize{(4.8)} & 48.3 \scriptsize{(6.2)} & 42.3 \scriptsize{(3.4)} \\         
         UniGCN \citep{ijcai21-UniGNN}  & 49.1 \scriptsize{(8.4)} & 74.4 \scriptsize{(3.9)} & 65.1 \scriptsize{(4.7)} & 51.3 \scriptsize{(6.3)} & 41.6 \scriptsize{(3.5)} \\
         UniGIN \citep{ijcai21-UniGNN} & 47.8 \scriptsize{(7.7)} & 69.8 \scriptsize{(5.6)} & 63.4 \scriptsize{(5.1)} & 48.3 \scriptsize{(6.1)} & 41.4 \scriptsize{(2.7)} \\
         UniGCNII \citep{ijcai21-UniGNN} & 48.5 \scriptsize{(7.4)} & 74.1 \scriptsize{(3.9)} & 65.8 \scriptsize{(3.9)} & 54.8 \scriptsize{(7.5)} & 42.5 \scriptsize{(3.9)} \\         
         AllSetTransformer \citep{chien2022you} & 47.6 \scriptsize{(4.2)} & 72.4 \scriptsize{(4.5)} & 65.3 \scriptsize{(3.9)} & 57.5 \scriptsize{(5.7)} & 42.3 \scriptsize{(2.4)} \\
         HyperGCL \citep{wei2022augmentations} & 60.3 \scriptsize{(7.4)} & 76.8 \scriptsize{(3.7)} & 79.7 \scriptsize{(3.8)} & 62.0 \scriptsize{(5.1)} & 43.9 \scriptsize{(3.6)} \\
         H-GD \citep{zheng2022rethinking} & 50.6 \scriptsize{(8.2)} & 74.5 \scriptsize{(3.5)} & 75.1 \scriptsize{(3.6)} & 58.8 \scriptsize{(6.2)} & 43.0 \scriptsize{(3.3)} \\
         ED-HNN \citep{wang2023equivariant} & 47.6 \scriptsize{(7.7)} & 72.7 \scriptsize{(4.7)} & 65.8 \scriptsize{(4.8)} & 54.8 \scriptsize{(5.4)} & 41.4 \scriptsize{(3.0)} \\
         HypeBoy \citep{kim2024hypeboy} & \underline{62.3} \scriptsize{(7.7)} & \underline{77.0} \scriptsize{(3.4)} & \underline{80.6} \scriptsize{(2.3)} & \underline{66.3} \scriptsize{(4.6)} & \underline{47.6} \scriptsize{(2.5)} \\
         \hline
         HyperBERT (Ours)  & \textbf{64.5 \scriptsize{(4.5)}} & \textbf{78.9 \scriptsize{(3.1)}} & \textbf{82.3 \scriptsize{(1.7)}} & \textbf{69.2 \scriptsize{(4.1)}} & \textbf{49.7 \scriptsize{(1.8)}} \\
        \hline
    \end{tabular}}
    }
    \label{tab:main_res}
\end{table*}

\section{Experiments}

In this section we show our model performance on five hypergraph node classification benchmarks from a variety of domains. Also, we present ablation studies to analyse the importance of the different components of the HyperBERT model.

\subsection{Datasets and Evaluation Metrics}

For our experiments we consider five datasets, with complete details included in Appendix \ref{appendix:datasets}. The hypergraph datasets are from diverse domains and sizes, expressing co-citation, co-authorship and movie-actor relations. Following previous works on hypergraph node classification \cite{wang2023equivariant,kim2024hypeboy}, we calculate classification accuracy in all our experiments. We evaluate all models over 10 runs and report the mean and standard deviation of the test set accuracy averaged across them.

\subsection{Overall Performance}

We first conducted experiments to evaluate model performance on text-attributed hypergraph node classification and present the results in Table \ref{tab:main_res}. As shown in the table, our model HyperBERT outperforms all the baselines on the five hypergraph evaluation datasets, so demonstrating superior capability in text-attributed hypergraph node classification. 

On the co-citation datasets, compared with AllSetTransformer \citep{chien2022you}, our model's classification accuracy is 64.5\% for the Cora dataset, achieving 16.9\% absolute performance improvement (47.6\% vs 64.5\%). In the case of PubMed, HyperBERT achieves 78.9\% classification accuracy, which when compared with AllSetTransformer (72.4\%) attains a 6.5\% absolute accuracy increment. When compared with more recent methods that make use of self-supervised techniques for training the model, such as HypeBoy \citep{kim2024hypeboy}, our model's classification accuracy is 64.5\% in Cora, a +2.2\% improvement from 62.3\%, and 78.9\% in PubMed, a +1.9\% increase from 77.0\%. This effectively shows the benefit of our proposed architecture for solving text-attributed hypergraph node classification tasks.

In the co-authorship dataset experiments, when compared with the AllSetTransformer, HyperBERT attains an accuracy of 82.3\% (+17\%) and 69.2\% (+11.7\%) in the DBLP-A and Cora-CA datasets, respectively. In comparison with HypeBoy, HyperBERT obtains +1.7\% and +2.9\% performance improvements in the DBLP-A and Cora-CA datasets, respectively. Furthermore, our model reaches a performance in the IMDB movies dataset of 49.7\%, an absolute improvement of 7.4\% and 2.1\% when compared with AllSetTransformer and HypeBoy, respectively.

\subsection{Ablation Study}

To investigate the contribution of each module in HyperBERT, we conduct several ablation studies to isolate and measure the impact of individual components on overall performance.

\paragraph{Effect of Architectural Components.} In order to validate the architectural design of HyperBERT, we explore the effect of its different components towards node classification accuracy in the Pubmed dataset, and provide the results in Table \ref{tab:ablation1}.  On the one hand, we investigate suppressing particular components of the loss function, such as the semantic loss, structural loss, and alignment loss. To do so, we set their respective $\lambda$ factors to 0, effectively cancelling their contribution during the pretraining stage. We find that these loss components lead to significant performance decreases when removed. Specifically, accuracy decreases to 66.4\% (-12.5\%), 68.8\% (-10.1\%) and 63.1\% (-15.8\%) when removing the semantic, structural and alignment losses, respectively. As expected, the biggest performance impact is given when eliminating the alignment loss, as it leads to completely disparate feature spaces for text and hypergraph node representations. We also study the impact of neglecting the high-order topology by replacing the HGNN encoder with a GNN, decreasing performance to 71.2\% (-7.7\%). In addition, we show the effect of not pretraining the model and instead training it from scratch on the downstream task, leading to a performance decrease of 71.5\% (-7.4\%). Therefore, the results of these experiments justify our architectural choices.

\begin{table}[h]
\centering
\resizebox{1\columnwidth}{!}{
\begin{tabular}{l|c}
\hline
Method     & Accuracy ($\%$) \\ 
\hline
HyperBERT w/o semantic loss & 66.4 $\pm$ 1.3 \\ 
HyperBERT w/o structural loss & 68.8 $\pm$ 1.6 \\ 
HyperBERT w/o alignment loss & 63.1 $\pm$ 2.3 \\ 
HyperBERT with GNN as structural layer & 71.2 $\pm$ 3.5 \\ 
HyperBERT w/o pretraining & 71.5 $\pm$ 1.5 \\ 
HyperBERT & \textbf{78.9 $\pm$ 3.1} \\ 
\hline
\end{tabular}}
\caption{Effect of the different architectural components in node classification accuracy (and $\pm$ 95\% confidence interval) of HyperBERT on the Pubmed benchmark.}
\label{tab:ablation1}
\end{table}

\paragraph{Impact of Hypergraph Encoder.} The choice of hypergraph encoder plays a crucial role in the computation of the hypergraph structural representations in HyperBERT. For this reason, Table \ref{tab:ablation2} shows the node classification results in the PubMed dataset when using several different hypergraph architectures as hypergraph encoder layers while keeping BERT as the text encoder. As can be observed, the highest performance is achieved when using HGNN as the hypergraph encoder, with 78.9\% ($\pm$ 3.1) accuracy, whereas the lowest is attained when using HyperGCN, leading to an accuracy of 76.3\% ($\pm$ 2.5). These results highlight the significant impact that the choice of hypergraph encoder can have on model performance, suggesting that careful selection of the encoder architecture is essential for optimizing node classification accuracy.

\begin{table}[h]
\centering
\resizebox{1\columnwidth}{!}{
\begin{tabular}{l|c}
\hline
Encoder     & Accuracy ($\%$) \\ 
\hline
HyperGCN \citep{10.5555/3454287.3454422} & 76.3 $\pm$ 2.5 \\ 
HNHN \citep{HNHN2020} & 76.9 $\pm$ 1.8 \\ 
UniGCN \citep{ijcai21-UniGNN} & 77.6 $\pm$ 2.3 \\ 
UniGIN \citep{ijcai21-UniGNN} & 77.5 $\pm$ 2.1 \\ 
UniGCNII \citep{ijcai21-UniGNN} & 77.9 $\pm$ 3.4 \\ 
ED-HNN \citep{wang2023equivariant} & 78.2 $\pm$ 2.6 \\ 
HGNN \citep{Feng_You_Zhang_Ji_Gao_2019} & \textbf{78.9 $\pm$ 3.1} \\ 
\hline
\end{tabular}}
\caption{Accuracy of node classification on the PubMed test set using different hypergraph architectures as hypergraph encoder layers.}
\label{tab:ablation2}
\end{table}

\paragraph{Impact of Text Encoder.} The choice of text encoder is crucial in determining the overall performance of HyperBERT, particularly when the hypergraph encoder is fixed to HGNN. Table \ref{tab:ablation3} presents the node classification results on the PubMed dataset using different text encoders while keeping HGNN as the hypergraph encoder. The highest performance is achieved with BERT, showing an accuracy of 78.9\% ($\pm$ 3.1), while ELECTRA and RoBERTa demonstrate slightly lower accuracies of 78.4\% ($\pm$ 2.9) and 78.2\% ($\pm$ 3.0), respectively. These results indicate that while all three text encoders provide competitive performance, BERT remains the most effective choice in this setup.

\begin{table}[h]
\centering
\resizebox{1\columnwidth}{!}{
\begin{tabular}{l|c}
\hline
Encoder     & Accuracy ($\%$) \\ 
\hline
RoBERTa \citep{liu2019roberta} & 78.2 $\pm$ 3.0 \\ 
ELECTRA \citep{clark2020electra} & 78.4 $\pm$ 2.9 \\ 
BERT \citep{Devlin2019BERTPO} & \textbf{78.9 $\pm$ 3.1} \\ 
\hline
\end{tabular}}
\caption{Accuracy of node classification on the PubMed test set using different text encoders while keeping HGNN as the hypergraph encoder.}
\label{tab:ablation3}
\end{table}

\section{Conclusion}

In this work, we propose a new model architecture to improve the capability of hypergraph structural encoding of BERT effectively while retaining its semantic encoding abilities. In order to achieve this, we designed HyperBERT, a hypergraph-aware language model based on BERT that augments with structural context for the challenging task of text-attributed hypergraph node classification. Notably, extensive experiments on five benchmark hypergraph datasets demonstrate the effectiveness of HyperBERT for text-attributed hypergraph node classification tasks, proving that simultaneously modelling hypergraph structure and semantic context is crucial to achieve state-of-the-art results. 

\section*{Limitations}

In this work we evaluate HyperBERT primarily on node classification tasks. However, there are other relevant downstream tasks, such as hyperedge prediction. Future work would benefit from investigating the applicability and capacity of HyperBERT to broader tasks.

\bibliography{hyperbert}

\clearpage
\newpage

\appendix

\section*{Appendix}

\section{Datasets}
\label{appendix:datasets}

For our experiments we use five publicly available benchmark hypergraph datasets. The hypergraph datasets are from diverse domains, expressing co-citation, co-authorship and movie-actor relations. Table \ref{tab:dataset_stats} provides statistics on the datasets. 

\begin{table}[h!]
\centering
\resizebox{0.9\columnwidth}{!}{
\begin{tabular}{l|ccc}
\hline
Dataset     & Nodes & Hyperedges & \# Classes \\ 
\hline
Cora & 1,434 & 1,579 & 7 \\ 
Pubmed & 3,840 & 7,963 & 3 \\ 
DBLP-A & 2,591 & 2,690 & 6 \\ 
Cora-CA & 2,388 & 1,072 & 7 \\ 
IMDB & 3,939 & 2,015 & 3 \\ 
\hline
\end{tabular}}
\caption{Statistics of the text-attributed hypergraph datasets used in our experiments.}
\label{tab:dataset_stats}
\end{table}

We utilize two co-citation datasets: Cora and Pubmed. In these datasets, each node represents a publication and hyperedges represent sets of publications co-cited by particular publications. For example, if a publication has cited other nodes (publications) $v_{i}$, $v_{j}$ and $v_{k}$, these nodes are grouped as a hyperedge \{$v_{i}$, $v_{j}$, $v_{k}$\} $\in$ $\mathcal{E}$. Node classes indicate categories of the publications. Moreover, for co-authorship datasets we utilize Cora-CA and DBLP-A. In Cora-CA and DBLP-A, each node represents a publication, and a set of publications that are written by a particular author is grouped as a hyperedge. Classes indicate categories of the publication. We use IMDB for a movie-actor dataset. In this dataset, each node indicates a movie, and the movies that a particular actor has acted on are grouped as a hyperedge. Node attributes describe the plot of the movie and classes indicate the genre of the movie.

\section{Implementation Details}
\label{appendix:implementation_details}

We implemented HyperBERT with PyTorch \cite{paszke_pytorch_2019} and the HuggingFace library \cite{wolf_transformers_2020}. For the hypergraph neural network components, we use PyTorch Geometric \cite{fey_fast_2019}. For the knowledge alignment stage, we set batch size as 32, and training steps to 20,000 on a single NVIDIA A100 GPU. We utilize the Adam optimizer with a fixed weight decay rate of $10^{-6}$ and learning rate of $10^{-3}$. The number of layers for HyperBERT is set to 6, number of heads is 8 and dropout rate is 0.5. The dimensionality is set to 512. The temperature parameter $\tau$ contained in our loss, is set to 0.2. For training on downstream tasks, we use 200 epochs with an output MLP for node classification containing 2 layers and dimensionality of 512. We evaluate validation accuracy of the model every 10 epochs and keep the model checkpoint that yields the best validation accuracy, then we use it for predicting on the test dataset. Results are on the test set unless stated otherwise.

\end{document}